\documentclass[11pt]{article}

\usepackage[final]{acl}

\usepackage{times}
\usepackage{latexsym}
\usepackage[utf8]{inputenc}
\usepackage[T1]{fontenc}
\usepackage{microtype}
\usepackage{amsmath}
\usepackage{amsfonts}
\usepackage{amssymb} 
\usepackage{nicefrac}
\usepackage{enumitem}
\usepackage{amssymb}
\usepackage{booktabs}
\usepackage{subcaption}
\usepackage{booktabs}
\usepackage{graphicx}
\usepackage{microtype}
\usepackage{marvosym}

\usepackage{url}                 

\usepackage{booktabs}
\usepackage{array}               
\usepackage{makecell}            
\usepackage{multirow}    
\usepackage{hyperref}

\usepackage{wrapfig}
\newcolumntype{C}[1]{>{\centering\arraybackslash}p{#1}}
\newcolumntype{P}[1]{>{\centering\arraybackslash}p{#1}}

\usepackage{graphicx}            
\usepackage[export]{adjustbox}   
\usepackage{wrapfig}             
\usepackage{pifont}         
\usepackage{xcolor}
\usepackage[font=small,tableposition=top]{caption}

\usepackage{inconsolata}

\usepackage{graphicx}

\newcommand{\ourbench}{Uni-MMMU}
\title{Uni-MMMU: A Massive Multi-discipline Multimodal Unified Benchmark}

\author{
Kai Zou\textsuperscript{1,3}\thanks{These authors contributed equally.},
Ziqi Huang\textsuperscript{2}\protect\footnotemark[1]\thanks{Project Leader},
Yuhao Dong\textsuperscript{2}\protect\footnotemark[1],
Shulin Tian\textsuperscript{2},
Dian Zheng\textsuperscript{4},\\
\textbf{Hongbo Liu\textsuperscript{1},
Jingwen He\textsuperscript{1,4},
Bin Liu\textsuperscript{3}\textsuperscript{\Letter},
Yu Qiao\textsuperscript{1}\textsuperscript{\Letter},
Ziwei Liu\textsuperscript{2}\textsuperscript{\Letter}}
\\[6pt]
\textsuperscript{1}Shanghai Artificial Intelligence Laboratory 
\textsuperscript{2}S-Lab, Nanyang Technological University \\
\textsuperscript{3}University of Science and Technology of China 
\textsuperscript{4}The Chinese University of Hong Kong\\
\\
\href{https://vchitect.github.io/Uni-MMMU-Project}{https://vchitect.github.io/Uni-MMMU-Project}
}

\begin{document}

\renewcommand{\thefootnote}{\fnsymbol{footnote}}
\maketitle
\setcounter{footnote}{0}
\renewcommand{\thefootnote}{\arabic{footnote}}

\begingroup
\renewcommand{\thefootnote}{\Letter}
\footnotetext{Corresponding authors.}
\endgroup

\begin{abstract}
Unified multimodal models aim to jointly enable visual understanding and generation, yet current benchmarks rarely examine their true integration. Existing evaluations either treat the two abilities in isolation or overlook tasks that inherently couple them. To address this gap, we present \textbf{Uni-MMMU}, a comprehensive and discipline-aware benchmark that systematically unfolds the bidirectional synergy between generation and understanding across eight reasoning-centric domains, including science, coding, mathematics, and puzzles. Each task is \textit{bidirectionally coupled}, demanding models to (i) leverage conceptual understanding to guide precise visual synthesis, or (ii) utilize generation as a cognitive scaffold for analytical reasoning. Uni-MMMU incorporates verifiable intermediate reasoning steps, unique ground truths, and a reproducible scoring protocol for both textual and visual outputs. Through extensive evaluation of state-of-the-art unified, generation-only, and understanding-only models, we reveal substantial performance disparities and cross-modal dependencies, offering new insights into \textit{when and how} these abilities reinforce one another, and establishing a reliable foundation for advancing unified models.
\end{abstract}
\section{Introduction}

Recent advances in large language models (LLMs)~\cite{floridi2020gpt} and high-fidelity image synthesis~\cite{rombach2022high, peebles2023scalable} have catalyzed multimodal systems that accept visual inputs and produce grounded, instruction-following outputs~\cite{liu2023visual, achiam2023gpt}. Building on this progress, \emph{unified} frameworks have emerged that aim to couple comprehension and generation within a single modeling or training recipe~\cite{deng2025emerging, wu2025omnigen2, gpt4o_systemcard_2024}. The motivation is clear: many real tasks interleave perception with synthesis; a shared representation promises tighter control loops, reduced orchestration overhead, and better data efficiency.

While unification is intuitively appealing, it remains unclear \emph{when} and \emph{how} generation (Gen) and understanding (Und) actually reinforce one another. Human cognition thrives on this synergy. To solve a challenging geometry problem, a student might draw auxiliary lines—a generative act—to create tangible visual cues that scaffold abstract deduction. Conversely, an artist painting a realistic scene leverages an understanding of optical principles to guide the generative act of depiction. Complex problem-solving thus involves an iterative loop between generation and understanding, where the correctness of both the intermediate steps and the final results is crucial for success and thus requires rigorous evaluation.

However, existing benchmarks for unified models fail to adequately assess this critical interaction. They are often limited in two ways: some evaluate generative or understanding capabilities in isolation~\cite{zhao2025envisioning}, while others that test both focus on superficial aspects like content association~\cite{chen2024interleaved} or cross-modal consistency~\cite{mollah2025telephone}. Crucially, they lack tasks that enforce a \textbf{necessary logical dependency} between the two processes—a dependency that is the cornerstone of complex, multi-step problem-solving.

To bridge this gap, we propose the Uni-MMMU, a novel suite of tasks designed to explicitly evaluate this mutual reinforcement. We curate eight tasks from logically rigorous disciplines such as mathematics, physics, and coding. These fields are ideal as they demand the same interplay of concrete visualization and abstract reasoning seen in human problem-solving. The Uni-MMMU framework probes two core synergistic pathways: \emph{Understanding aids Generation} and \emph{Generation aids Understanding}. Each task is built upon a foundational design of a deterministic reasoning path and a unique correct answer, enabling a dual-level assessment of both the final outcome and intermediate steps. Crucially, this structure underpins our fully automated and reproducible evaluation pipeline, which employs a combination of programmatic parsers, perceptual metrics, and LLM-as-a-Judge to deliver objective, consistent, and interpretable results across all models.

Our evaluation using Uni-MMMU on state-of-the-art models yields a critical insight: the synergy between generation and understanding is most potent in tasks with strong logical dependencies. In these scenarios, intermediate modal information is pivotal, even imperfect, model-generated intermediates significantly improve final accuracy over end-to-end approaches, while oracle intermediates lead to substantial performance gains. This analysis also reveals a clear imbalance in current unified models: they are heavily skewed towards understanding, with generation acting as the primary bottleneck. Common failure points include imprecise image editing, the synthesis of schematic diagrams, and fine-grained spatial reasoning.

In summary, our main contributions are as follows:
\begin{itemize}
  \item We propose \textbf{Uni-MMMU}, a benchmark of eight \emph{bidirectionally coupled} tasks that enforce Gen–Und logical dependency.
  \item We design a \emph{deterministic, dual-level} evaluation protocol with oracle trajectories and programmatic metrics for reliable, interpretable, and reproducible assessment.
  \item We provide a comprehensive, multi-discipline evaluation of unified and specialized models, diagnosing where synergy holds and where it breaks.
\end{itemize}

\section{Related Works}

\begin{figure*}[ht]
\begin{center}
\includegraphics[width=1.0\textwidth]{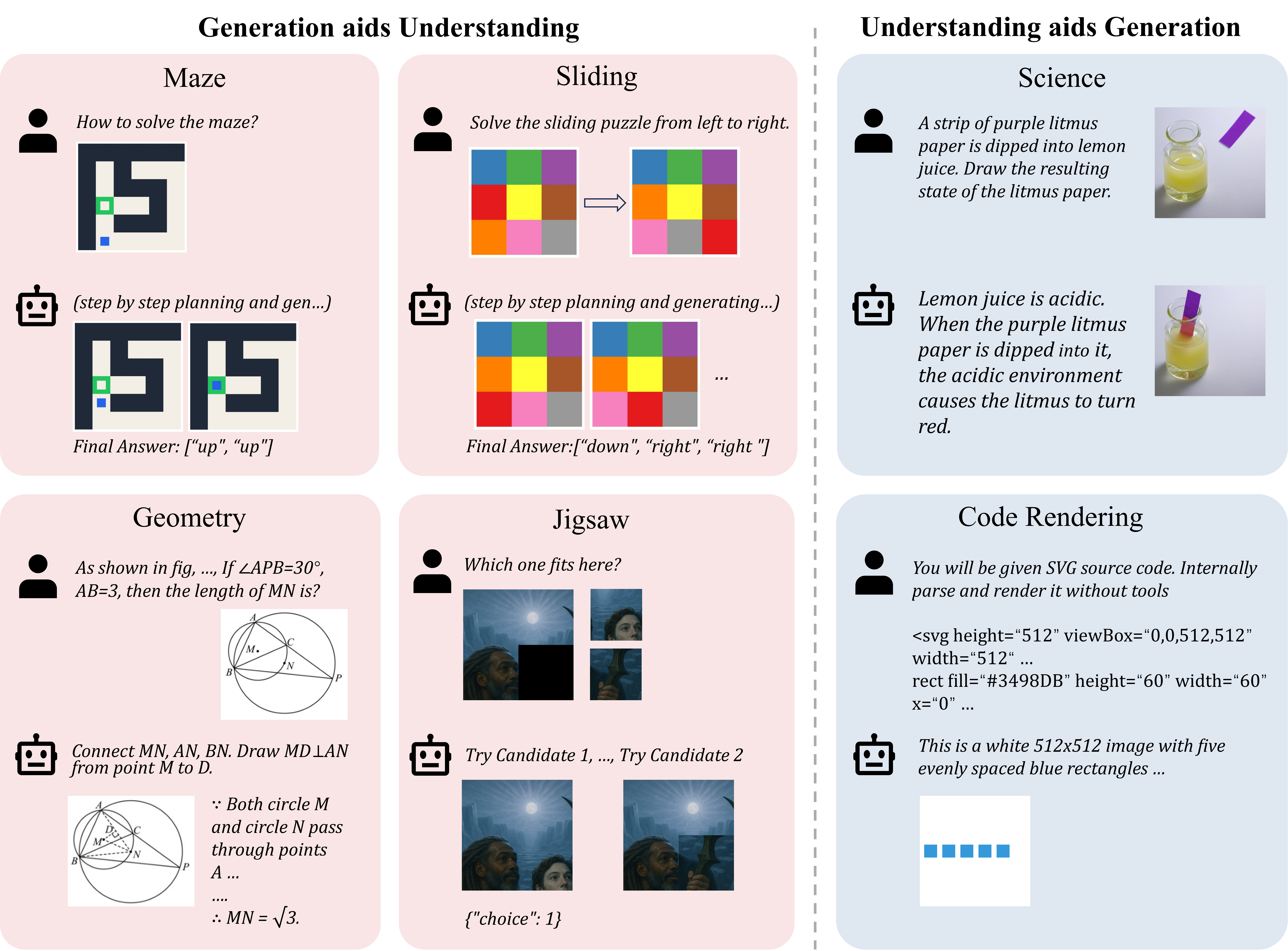}
\end{center}
\caption{\textbf{Overview of Uni-MMMU.} Eight tasks are grouped into two paradigms: generation aids understanding (Maze, Sliding, Geometry, Jigsaw) and understanding guides generation (Science: Physics/Chemistry/Biology; Code Rendering). Each task reports dual-channel scores (text + image).}
\label{fig:overview}
\end{figure*}

\subsection{Unified Multimodal Models}
Recently, there has been growing interest in developing unified models that integrate both understanding and generation capabilities. Emu3~\cite{wang2024emu3} primarily focused on directly combining understanding and generation modules, whereas VILA-U~\cite{wu2024vila, chen2025janus} has explored systematic integration strategies that enable these abilities to complement each other rather than merely merging them. Among these efforts, BAGEL~\cite{deng2025emergingpropertiesunifiedmultimodal} established one of the first widely adopted baselines by employing mixed transformer experts to realize emerging unified multimodal intelligence. Show-o2~\cite{xie2025showo2improvednativeunified} further supports video generation within a unified modeling framework. OneCAT~\cite{li2025onecatdecoderonlyautoregressivemodel} proposed a decoder-only architecture that achieves native multimodal unification, while UniPic~\cite{wei2025skyworkunipic20building} introduced reinforcement learning algorithms to further advance performance. Despite these advances, most existing models are still evaluated separately on understanding and generation benchmarks, revealing the lack of a comprehensive benchmark for assessing their joint capabilities from a unified perspective. To address this gap, we propose Uni-MMMU, a holistic benchmark designed to evaluate unified models in an integrated manner and to highlight the synergy between their understanding and generation abilities.

\newcommand{\cmark}{\textcolor{green!70!black}{\ding{51}}}%
\newcommand{\xmark}{\textcolor{red}{\ding{55}}}%

\begin{table}[t!]
\centering
\resizebox{\linewidth}{!}{%
\begin{tabular}{lcccc}
\toprule
                        & \textbf{MMU} & \textbf{Gen\&Edit} & \textbf{Multi-Turn} & \textbf{Dual Eval} \\ \hline
MMMU                    & \cmark       & \xmark             & \xmark              & \xmark             \\
WISE                    & \xmark       & \xmark             & \xmark              & \xmark             \\
RISEBench               & \xmark       & \cmark             & \xmark              & \xmark             \\
OpenING                 & \cmark       & \cmark             & \cmark              & \xmark             \\
MME-Unify               & \cmark       & \cmark             & \xmark              & \xmark             \\
UniEval                 & \cmark       & \xmark             & \xmark              & \xmark             \\
\textbf{Uni-MMMU}    & \cmark       & \cmark             & \cmark              & \cmark             \\ \hline
\end{tabular}}
\caption{
    \textbf{Uni-MMMU VS. prior benchmarks}. MMMU~\cite{yue2024mmmu}, WISE~\cite{niu2025wise}, RISEBench~\cite{zhao2025envisioning}, MME-Unify~\cite{xie2025mme}, UniEval~\cite{li2025unieval}, and OpenING~\cite{zhou2025opening}. 
    We compare across four key dimensions: multimodal understanding (\textbf{MMU}), generation and editing (\textbf{Gen\&Edit}), multi-turn evaluation (\textbf{Multi-Turn}), and dual evaluation of the process and result (\textbf{Dual Eval}).
}
\label{tab:Comparison}
\end{table}

\subsection{Multimodal Understanding and Generation Benchmarks}

Evaluation benchmarks in multimodal AI are evolving, moving beyond siloed assessments of individual capabilities. In understanding, the focus has shifted from basic perception, as in MMMU~\cite{yue2024mmmu}, toward integrated "thinking-with-images" paradigms that incorporate generation~\cite{su2025thinkingimagesmultimodalreasoning}. Similarly, generation evaluation has progressed from measuring basic semantic fidelity to assessing complex, understanding-driven tasks in frameworks such as ImgEdit~\cite{ye2025imgeditunifiedimageediting} and Understanding-in-Generation~\cite{lyu2025understandingingenerationreinforcinggenerativecapability}. This evolution extends to new evaluation frameworks like Evaluation-Agent~\cite{zhang2024evaluationagent} and in the video domain~\cite{huang2023vbench, huang2024vbench++, zheng2025vbench2}.

While this reflects a clear trend toward synergistic models, a critical gap persists in evaluation: existing unified benchmarks like MME-Unify~\cite{xie2025mme} still assess understanding and generation largely in isolation, failing to probe their interplay. To bridge this gap, we propose Uni-MMMU, a benchmark designed to directly assess this interaction. As summarized in Tab.~\ref{tab:Comparison}, Uni-MMMU provides a more comprehensive assessment by uniquely combining multimodal understanding with generation, incorporating complex multi-turn tasks, and crucially, performing a dual evaluation of both the intermediate process and the final result to enable fine-grained error attribution.

\section{The Uni-MMMU}

As shown in Fig.~\ref{fig:overview}, \ourbench{} is a comprehensive benchmark designed to evaluate the capabilities of unified multimodal models across diverse tasks that require integrated understanding and generation abilities. Our benchmark systematically assesses models' capacity to leverage visual generation as an auxiliary tool for enhanced reasoning and problem-solving.

\subsection{Uni-MMMU Benchmark Suite}

\begin{figure}[t]
\begin{center}
\includegraphics[width=.9\linewidth]{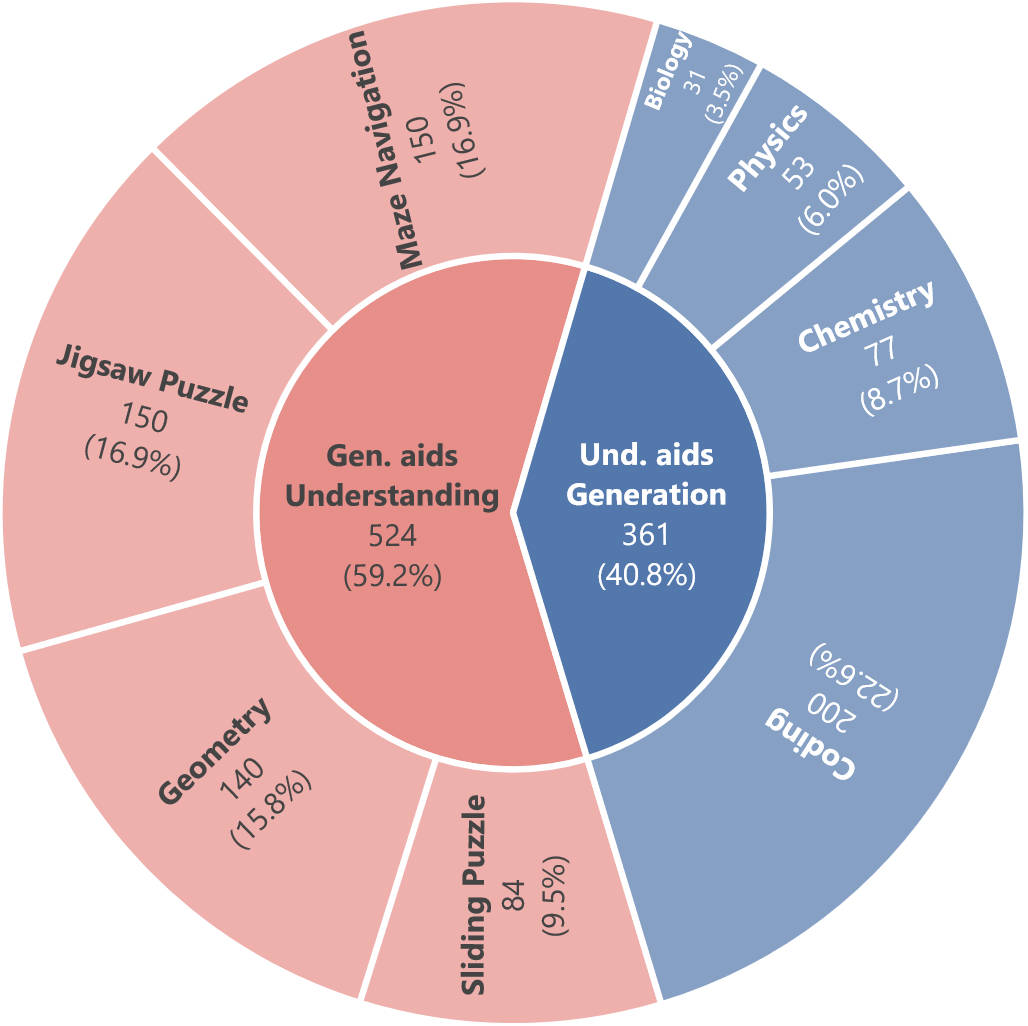}
\end{center}
\caption{\textbf{Data distribution in Uni-MMMU.} The chart illustrates the breakdown of the 885 instances into two primary paradigms—Generation aids Understanding (59.2\%) and Understanding aids Generation (40.8\%)—and further details the distribution across eight distinct disciplines.}
\label{fig:bin}
\end{figure}

To better evaluate the unified models' capabilities of mutual enhancement for generation and understanding tasks, we curate a diverse collection of tasks organized into distinct categories based on the type of vision-language interaction required as shown in Fig.~\ref{fig:bin}.
\subsubsection{Generation aids Understanding} 
This paradigm focuses on tasks where \textit{visual generation serves as an external cognitive scaffold} to support intermediate reasoning steps. Rather than treating visual outputs as final products, these tasks require models to iteratively generate visual states and use them to solve complex problems, mirroring how humans use sketches, diagrams, or step-wise visualizations for spatial reasoning.

\paragraph{Maze Navigation.} This task challenges a model's visual state tracking and pathfinding. Models receive a $6 \times 6$ ``perfect maze'' image as input, with a blue block marking the start, a green frame marking the goal, black walls, and white paths. The model must plan and execute the unique shortest path to the goal, alternately generating (1) the next move direction and (2) the corresponding updated maze image, ultimately outputting the full move sequence as text.
Mazes are procedurally generated using \textit{DFS carving} to ensure a loop-free structure and \textit{BFS verification} to guarantee a unique solution. Only mazes with shortest path lengths between 2 and 10 are retained.

\paragraph{Sliding Puzzle.} This task evaluates optimal state-space search and visual execution. Models are given the initial and goal states of a $3 \times 3$ 8-puzzle rendered in a fixed 9-color palette. The task is to produce the shortest solution sequence, alternating between textual move descriptions and visual puzzle states after each move, culminating in a JSON list of moves. 
Puzzles are generated by applying random moves to the solved state, then using BFS to verify solvability and uniqueness of the optimal path. Instances with multiple shortest paths are discarded.

\paragraph{Geometry with Auxiliary Lines.} 
This task directly tests a model's ability to scaffold logical deduction with visual constructs. Models are required to solve geometry problems by first interpreting textual instructions to generate a new diagram with correctly drawn auxiliary lines. Using this self-generated figure as a visual aid, they must then produce a step-by-step textual solution. The problems are sourced from the Geo-Laux benchmark~\citep{fu2025geolaux} and come complete with the original figures, line instructions, and ground-truth diagrams.

\paragraph{Jigsaw Puzzle.} 
This task assesses visual coherence through conditional generation and comparative reasoning. Models are given a $2 \times 2$ image panel with one missing patch and two potential candidates. The required process involves two stages: first, the model must sequentially generate two completed images, one for each candidate patch. Second, it must reason over its own generated outputs to textually justify and decide which candidate correctly completes the panel. Each instance is created by cropping a $3 \times 3$ grid from a sample from ShareGPT-4o-Image~\citep{chen2025sharegpt} dataset, masking the bottom-right patch, and selecting a distractor from the remaining patches.

Across these tasks, visual generation functions as an external working memory: models must maintain and update structured spatial states over multiple steps, which would be error-prone if done purely textually. Each generated visual serves both as a \textit{validation checkpoint} and as a \textit{planning canvas}, enabling robust multi-step spatial reasoning.

\subsubsection{Understanding aids Generation} 
This paradigm flips the evaluation direction: models must first understand and reason, then generate images based on their reasoning, assessing whether they can use conceptual understanding to guide structured generation.

\paragraph{Physics.}
The tasks probe a model's qualitative reasoning on core physical principles across mechanics, thermodynamics, and electromagnetism. Scenarios are designed to have unambiguous and deterministic visual outcomes, such as predicting changes due to thermal expansion or magnetic attraction. We construct these tasks using a hierarchical, LLM-driven pipeline to procedurally generate textual prompts for initial and final states. These prompts guide the synthesis of corresponding images, which are then manually curated to ensure scientific validity.

\paragraph{Chemistry.}
The tasks test a model's capacity to predict the visual manifestations of common chemical reactions. The content covers acid-base reactions, oxidation, and precipitation, with scenarios selected specifically for their salient visual cues, like distinct color changes or the formation of a solid. This design ensures that the reasoning is directly linked to the observable outcomes of fundamental chemical principles, such as predicting the result of mixing two reactive solutions.

\paragraph{Biology.}
The tasks target a model's reasoning about fundamental life science processes and their macroscopic consequences. The content focuses on plant physiology and cellular phenomena, featuring scenarios with clear, observable changes over time. Key concepts include phototropism, fruit ripening, and osmosis, which require the model to visualize the physical transformation resulting from an underlying biological mechanism, such as a plant bending towards light.

\paragraph{Code Rendering.}
The Code Rendering task probes a model's ability to natively interpret and visualize programmatic graphics. Models are given raw SVG snippets and must (i) produce a concise natural-language description of the depicted scene and (ii) render an image faithful to the SVG specification. The SVG corpus is procedurally generated at three difficulty tiers: \emph{simple} (single primitive), \emph{medium} (multiple non-overlapping primitives), and \emph{complex} (overlapping geometry, control-flow constructs, and symbol reuse).

\begin{table*}[t!]
\centering
\resizebox{\textwidth}{!}{%
\begin{tabular}{lccccccccccccccc}
\toprule
\multicolumn{1}{c}{} &
\multicolumn{8}{c}{\textbf{Generation aids Understanding}} &
\multicolumn{6}{c}{\textbf{Understanding aids Generation}} &
\multicolumn{1}{c}{} \\
\cmidrule(lr){2-9} \cmidrule(lr){10-15}
\textbf{Model} &
\textbf{Jig. I} & \textbf{Jig. T} & \textbf{Maze I} & \textbf{Maze T} & \textbf{Slid. I} & \textbf{Slid. T} & \textbf{Geo I} & \textbf{Geo T} &
\textbf{Sci. R.} & \textbf{Sci. T} & \textbf{Sci. I} & \textbf{C. T} & \textbf{C. S} & \textbf{C. P} &
\textbf{Avg.} \\
\midrule
\multicolumn{16}{c}{\textit{Unified Models}} \\
\midrule
Bagel            & 56.0 & 48.0 & 0.0/0.0 & 0.0/0.0 & 0.0/0.0 & 5.6/1.2 & 8.5  & 32.8 & 63.1 & 57.3 & 28.0 & 53.0 & 2.2 & 1.8 & 22.0 \\
OmniGen2         & 70.3 & 48.0 & 0.0/0.0 & 0.0/0.0 & 0.0/0.0 & 0.0/0.0 & 4.2  & 5.7  & 42.0 & 33.1 & 8.9  & 17.0 & 13.2 & 13.3 & 16.0 \\
Ovis-U1          & 57.0 & 53.0 & 0.0/0.0 & 12.5/0   & 0.0/0.0 & 0.0/0.0 & 7.1  & 3.5  & 42.7 & 36.3 & 24.8 & 18.0 & 8.0  & 10.5 & 16.5 \\
Qwen-Image-Edit  & 72.0 & 43.3 & 0.0/0.0 & 13.8/0.7 & 0.0/0.0 & 3.6/0.0  & 12.8 & 8.5  & 61.1 & 50.3 & 26.7 & 36.0 & 23.5 & 20.3 & 26.3 \\
nano-banana     & 48.9 & 57.0 & 1.8/0.0 & 23.3/4.7 & 1.0/0.0 & 6.2/0.0  & 21.4 & 47.8 & 91.7 & 79.6 & 43.9 & 75.0 & 36.5 & 33.7 & 37.3 \\
GPT4.1 + GPT-image              & 80.7 & 80.0 & 0.8/0.7 & 49.0/18.1& 8.4/0.0 & 25.6/2.4 & 25.7 & 17.1 & 93.6 & 91.1 & 61.8 & 71.6 & 83.6 & 68.6 & 44.1 \\
\midrule
\multicolumn{16}{c}{\textit{Generation-only}} \\
\midrule
FLUX.1-Kontext   & \multicolumn{8}{c}{\textit{-}} &  -- & -- & 9.5 & -- & 0.1 & 0.1 & -- \\
Imagen 3-001     & \multicolumn{8}{c}{\textit{-}} &  -- & -- & 8.3 & -- & 0.2 & 0.3 & -- \\
Imagen 4         & \multicolumn{8}{c}{\textit{-}} &  -- & -- & --  & -- & 54.8 & 51.6 & -- \\
\midrule
\multicolumn{16}{c}{\textit{Understanding-only}} \\
\midrule
Qwen2.5-VL-72B   &  --  & 72.6 &  -- & 42.1/8.7 & -- & 33.0/2.4 & -- & 18.6 & \multicolumn{6}{c}{\textit{-}} & -- \\
GPT4.1           &  --  & 78.0 &  -- & 56.1/9.4 & -- & 24.0/2.4 & -- & 16.4 & \multicolumn{6}{c}{\textit{-}} & -- \\
Gemini-2.5 Pro    &  --  & 71.3 &  -- & 51.3/8.7 & -- & 44.7/39.3& -- & 52.1 & \multicolumn{6}{c}{\textit{-}} & -- \\
\bottomrule
\end{tabular}%
}
\caption{
    \textbf{Model Performance Comparison on the Uni-MMMU Benchmark.} 
    All scores are normalized to a \textbf{[0, 100]} scale for consistency. 
    \textbf{Column abbreviations} are as follows: 
    \textbf{Jig.} (Jigsaw), 
    \textbf{Maze} (Maze Navigation), 
    \textbf{Slid.} (Sliding Puzzle), 
    \textbf{Geo} (Geometry), 
    \textbf{Sci.} (Science), and 
    \textbf{C.} (Code Rendering). 
    \textbf{I} (Image accuracy), 
    \textbf{T} (Text accuracy), 
    \textbf{R} (Reasoning accuracy), 
    \textbf{S} (Shape\&Color score), and 
    \textbf{P} (Position score). 
    For multi-step tasks (Maze, Sliding Puzzle), scores in the format \textit{a/b} represent \textit{step-level accuracy / sample-level accuracy}. 
}
\label{tab:model_performance}
\end{table*}

\subsection{Multi-discipline Evaluation Method Suite}
\label{sec:evaluation_pipeline}

The evaluation of \ourbench{} is underpinned by a fine-grained, programmatic pipeline designed to jointly assess both textual and visual outputs. To ensure objectivity and reproducibility over subjective human scoring, we employ a combination of \textit{(1) deterministic parsers}, \textit{(2) perceptual similarity metrics}, and \textit{(3) model-as-a-judge evaluations}. While each task category features a tailored protocol, a shared philosophy guides our approach: we meticulously evaluate the intermediate visual steps in a process and separately assess the final textual outcome against ground-truth solutions.

\paragraph{Maze Navigation \& Sliding Puzzle.}
For tasks demanding \textbf{multi-step spatial planning}, we score both the intermediate visual states and the final textual answers:
\begin{itemize}[leftmargin=10pt,topsep=0pt]
    \item \textbf{Image Evaluation.} Each generated state image is processed by a deterministic color-discretization parser. For mazes, the parser crops the $6 \times 6$ region and classifies each cell via pixel color distance against a fixed palette, using a $75\%$ majority threshold. For sliding puzzles, it classifies each $3 \times 3$ cell by its dominant color with an $80\%$ tolerance. We report two metrics:
    \begin{itemize}
        \item \texttt{img\_sample\_acc}: A binary score, $1$ if \emph{all} generated images in a sequence perfectly match the ground-truth grids after parsing.
        \item \texttt{img\_step\_acc}: The fraction of correctly parsed images, averaged across all steps in the solution.
    \end{itemize}

    \item \textbf{Text Evaluation.} The final predicted action sequence (\emph{e.g.}, movement directions or sliding operations) is parsed from the model's text output and compared against the ground-truth path:
    \begin{itemize}
        \item \texttt{text\_sample\_acc}: A binary score, $1$ if the entire predicted sequence exactly matches the ground truth.
        \item \texttt{text\_step\_acc}: The proportion of correctly matched moves, evaluated position-wise.
    \end{itemize}
\end{itemize}

This dual-pronged evaluation allows us to precisely diagnose failures, determining whether they stem from inaccurate state representation (visual errors) or flawed reasoning over those states (textual errors).

\paragraph{Jigsaw Puzzle.} 
Evaluation for jigsaw tasks is twofold, covering both image reconstruction quality and final decision correctness:
\begin{itemize}[leftmargin=10pt,topsep=0pt]
    \item \textbf{Image Quality.} 
    For each candidate patch, the model must generate two completed $2 \times 2$ panels. Their perceptual similarity to the corresponding ground-truth composites is measured using the DreamSim~\cite{fu2023dreamsim} metric, which jointly considers low-level structure and high-level semantics. 
    
    \item \textbf{Decision Accuracy.} 
    The model's final structured JSON output is parsed to extract its chosen candidate index (0 or 1), which is compared to the ground-truth label to compute \texttt{text\_sample\_acc}.
\end{itemize}

\paragraph{Geometry with Auxiliary Lines.} 
This evaluation employs a \textbf{model-as-a-judge} framework using strong open-weights Vision-Language Models (see Tab.~\ref{tab:agreement} for validation):

\begin{itemize}[leftmargin=10pt,topsep=0pt]
    \item \textbf{Image Accuracy} (\texttt{image\_acc}).
    The generated diagram with auxiliary lines is evaluated by the open-weights Qwen2.5-VL-72B~\citep{bai2025qwen2_5_vl}, which receives (1) the original figure, (2) the textual auxiliary line instructions, (3) the ground-truth auxiliary line figure, and (4) the model's generated figure. It returns a binary correctness score, tolerating minor stylistic differences but not geometric errors.
    
    \item \textbf{Text Accuracy} (\texttt{text\_acc}). 
    The model's step-by-step textual solution is judged by the open-weights Qwen3-32B~\citep{yang2025qwen3}, which checks both the logical rigor of the reasoning and the correctness of the final numerical or symbolic answer.
\end{itemize}

\paragraph{Scientific Reasoning (Physics, Chemistry, and Biology).}
Evaluation for these tasks is conducted by a VLM judge, which assesses the model's output across three dimensions. It first examines the textual output on two criteria: the correctness of the scientific reasoning (\texttt{text\_reason\_acc}), ensuring the explanation applies relevant principles, and the physical accuracy of the described outcome (\texttt{text\_result\_acc}). The judge then evaluates the generated image (\texttt{img\_acc}), verifying that it visually and semantically corresponds to the ground-truth final state while maintaining consistency with the initial scene.

\paragraph{Code Rendering.}
A VLM judge evaluates the outputs using a fine-grained qualitative rubric that compares against the ground-truth rendering. It first assesses the model-generated textual summary (\texttt{text\_acc}) for semantic consistency, checking for correct object types, counts, colors, and relative layout. The judge then scores the rendered image itself along two axes: geometric and color fidelity (\texttt{shape\_color\_acc}), which covers the correctness of shapes (including polygon side counts) and color usage on a 0--5 scale; and spatial precision (\texttt{position\_acc}), which evaluates the accuracy of the object layout, alignment, spacing, rotation, and relative placement within the canvas.

A key design choice is that both text and image channels are evaluated separately and jointly, ensuring that a model cannot compensate poor reasoning with good rendering or vice versa. This dual-modality evaluation allows us to isolate whether failures stem from perception, generation, reasoning, or the integration thereof. We deliberately select open-weights models as judges for long-term accessibility and reproducibility; their reliability is validated in Tab.~\ref{tab:agreement}.

\section{Experiments}
\subsection{Experiment Setup}

We evaluated a suite of 6 advanced unified models and 6 specialized models. The unified models include four leading open-weights systems: Bagel~\cite{deng2025emerging}, OmniGen2~\cite{wu2025omnigen2}, Ovis-U1~\cite{ovisU1}, and Qwen-Image-Edit~\cite{qwen_image_2025}; nano-banana (gemini-2.5-flash-image), which is capable of automatic interleaved image and text generation; and GPT (GPT4.1 + GPT-image)~\cite{gpt4o_systemcard_2024}, a closed-source agent-based model where GPT4.1 invokes GPT-image for image synthesis. With the exception of nano-banana, all other unified models required manual iterative calls to achieve interleaved generation. For Ovis-U1, which accepts only a single image as a VAE-encoded reference, we used the output image from the previous step for the Maze and Sliding tasks, and a stitched concatenation of the input images for the Jigsaw task. Our evaluation also includes specialized generative models such as the Diffusion Transformer-based FLUX.1-Kontext-dev~\cite{batifol2025flux1}, Imagen 3-001~\cite{saharia2022imagen}, and Imagen 4, and leading-edge Large Vision-Language Models (LVLMs) for understanding, including Qwen2.5-VL-72B~\cite{bai2025qwen2_5_vl}, GPT4.1~\cite{gpt4o_systemcard_2024}, and Gemini-2.5 Pro~\cite{gemini_2_5_report_2025}.

\begin{table}[t!]
\centering
\resizebox{.9\linewidth}{!}{%
\begin{tabular}{lcc}
\hline
                      &  $\kappa_{img}$ & $\kappa_{txt}$     \\ \hline
Ours vs Gemini-2.5-pro & 0.7512                & 0.7538          \\
Ours vs Human          & 0.717                 & 0.7404          \\ \hline
\end{tabular}}
\caption{\textbf{Validity of Uni-MMMU.} The table shows the inter-rater reliability, measured by Cohen's Kappa ($\kappa$), between our model-as-a-judge evaluator and both a stronger proprietary model (Gemini-2.5-pro) and professional human annotators.}
\label{tab:agreement}
\end{table}

\subsection{Uni-MMMU Evaluation Results}

\noindent\textbf{Generation aids Understanding.}
As shown on the left side of the Tab.~\ref{tab:model_performance}, we present the performance of various models on the Maze, Sliding, Jigsaw, and Geometry tasks, with all metrics normalized to a 0-100 scale. On the Jigsaw task, GPT achieves the highest scores in both image generation and textual understanding, whereas Bagel performs the lowest in both categories. This suggests a positive correlation between image generation quality and final reasoning accuracy. This finding is further corroborated in the Maze and Sliding tasks, where GPT and nano-banana leverage their superior image generation capabilities to aid reasoning, thereby achieving better path-planning performance. (Nano-banana's lower score on Jigsaw is largely attributable to the additional difficulty imposed by its automatic generation capability, which can lead to penalties for producing an incorrect number of images.) Among the open-source models, Bagel demonstrates strong reasoning abilities, evidenced by its leading final score on the Geometry task. In comparison to specialized understanding models, we observe that unified models generally score lower on Jigsaw. This is primarily because the intermediate generative steps create longer and more complex contexts, which degrades their ability to adhere to the specific formatting instructions for the final answer. Gemini-2.5-pro exhibits the strongest reasoning capabilities, demonstrated by its top performance in Geometry problem-solving and its exceptionally advanced spatial planning skills on the less complex Sliding task.

\noindent\textbf{Understanding aids Generation.}
As shown on the right side of the Tab.~\ref{tab:model_performance}, we present the performance of different models on the Science and Code tasks, with all metrics also normalized to a 0-100 scale. On the Science task, a strong positive correlation is evident between understanding/reasoning ability and final generation quality. OmniGen2 exhibits the weakest reasoning and image generation abilities, whereas GPT achieves the highest generation accuracy, attributable to its superior world knowledge and understanding. This observation is reinforced by the performance of non-understanding models (e.g., FLUX-Kontext), which also show the lowest image accuracy. The Code task reveals a notable finding: despite Bagel's strong understanding capabilities, it achieves low image accuracy as it is not adept at producing simple graphical outputs. Furthermore, Imagen4's high score on this task unveils its strong underlying understanding and reasoning abilities. A general trend across models is that positional accuracy is consistently lower than shape accuracy, indicating a common deficiency in precise spatial awareness.

\begin{table}[t]
\centering

\resizebox{\linewidth}{!}{%
\begin{tabular}{lccccccc}
\toprule
\textbf{Model} & \textbf{Jig.} & \textbf{Maze} & \textbf{Sliding} & \textbf{Math} & \textbf{Sci.} & \textbf{Code S} & \textbf{Code P} \\
\midrule
GPT-image & - & - & - & - & 55.4 & 81.5 & 66.2 \\
GPT-4.1 & 78.0 & 56.1/9.4 & 24.0/2.4 & 16.4 & - & - & - \\
\midrule
GPT & 80.0 & 49.0/18.1 & 25.6/2.4 & 17.1 & 61.8 & 83.6 & 68.6 \\
GPT w/ GT & 98.0 & 68.1/24.8 & 34.7/3.6 & 27.8 & 84.0 & 83.7 & 72.6 \\
\bottomrule
\end{tabular}%
}
\caption{
    \textbf{Ablation study.} The table compares the performance of the understanding-only module (GPT-4.1), the unified model generating its own intermediate steps (GPT), and an oracle setup with ground-truth intermediates (GPT w/ GT).
}
\label{tab:ab}

\end{table}

\subsection{Validity of Uni-MMMU}

To validate the efficacy of the LLM-based evaluation components in our methodology, we sampled 150 model outputs from across the Math, Science, and Code tasks. These samples were independently scored by professional human annotators and by a more powerful commercial model, Gemini-2.5-pro, using the identical scoring rubrics as our LLM judge. To facilitate a binary comparison for the Code task, we converted the multi-point image scores into a binary metric: an image was considered a positive sample if both its `Shape\&Color' and `Position' scores were greater than or equal to 4, and a negative sample otherwise. As shown in Tab.~\ref{tab:agreement}, we report the Cohen's Kappa coefficient, a metric that measures inter-rater agreement while correcting for the possibility of agreement occurring by chance and is robust to imbalanced class distributions. The evaluator designed in our method demonstrates a Cohen's Kappa coefficient in the 0.6 to 0.8 range when compared with both human evaluators and Gemini-2.5-pro, indicating substantial agreement. Notably, the consistency for textual evaluations was even higher, demonstrating the strong discriminative performance in assessing understanding capabilities. To complement the binary Cohen's Kappa with a more fine-grained measure, we additionally computed the Pearson correlation on the full 6-level scale of the Code Rendering task (707 samples), yielding coefficients of $0.84$ for Shape\&Color and $0.76$ for Position, further confirming strong agreement between the VLM judge and human annotators.

\begin{figure}[t]
  \centering

  \begin{subfigure}[t]{\linewidth}
    \centering
    \includegraphics[width=\linewidth]{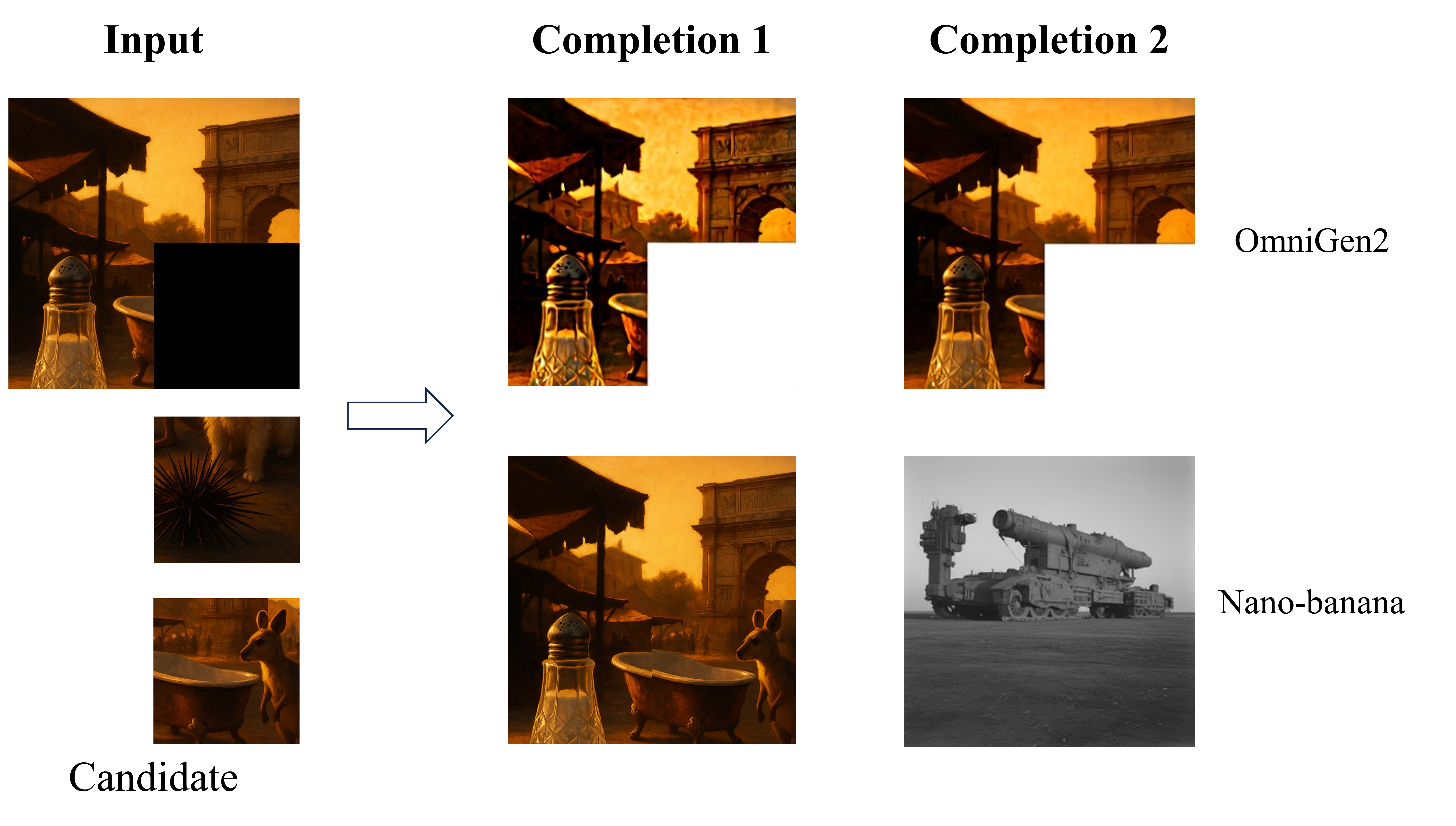}
    \caption{Error examples for Jigsaw.}
    \label{fig:error_jig}
  \end{subfigure}

  \vspace{0.6em}

  \begin{subfigure}[t]{\linewidth}
    \centering
    \includegraphics[width=\linewidth]{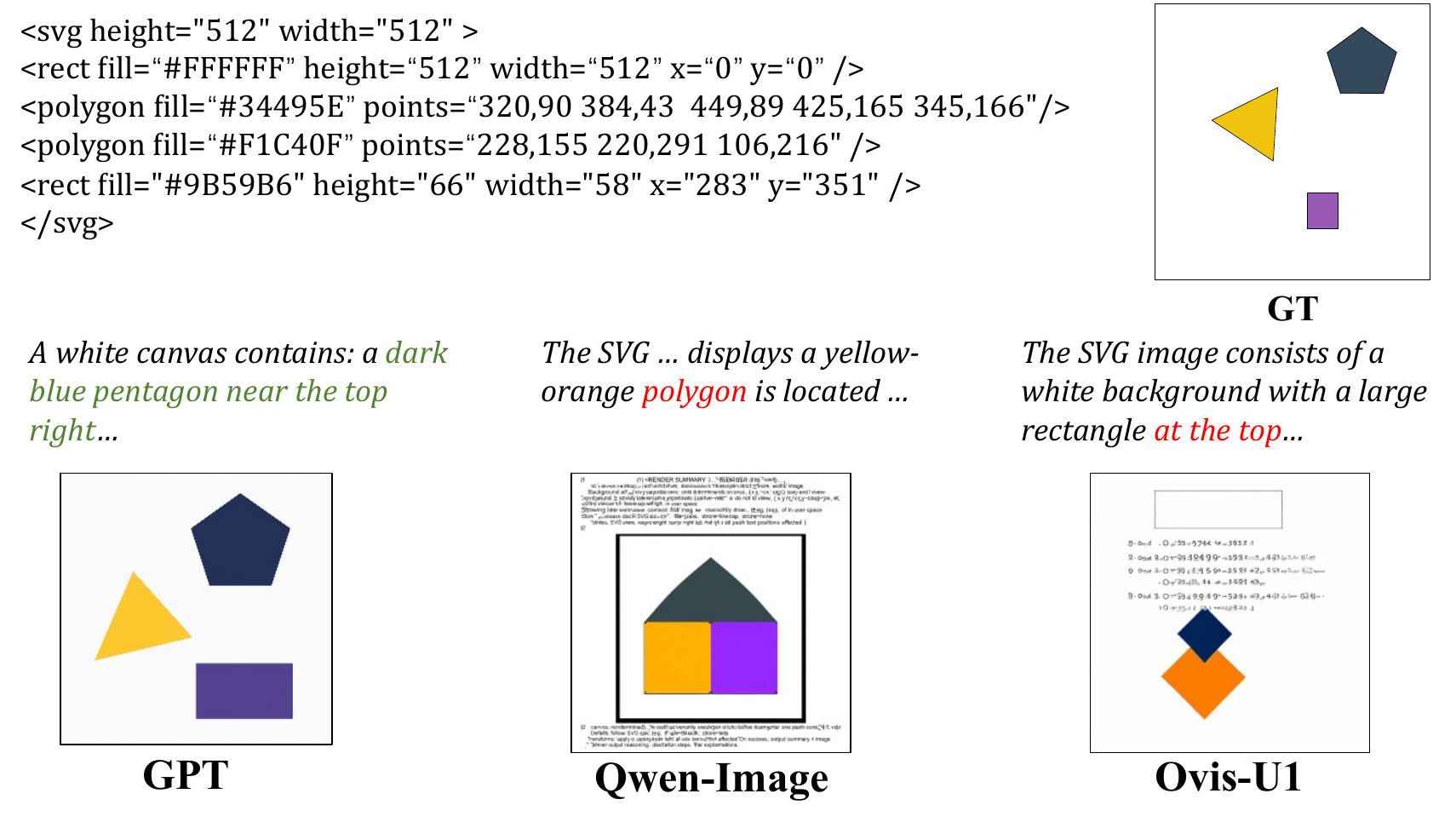}
    \caption{Examples of Code Rendering.}
    \label{fig:error_svg}
  \end{subfigure}

  \vspace{1.0em}

  \begin{subfigure}[t]{\linewidth}
    \centering
    \includegraphics[width=\linewidth]{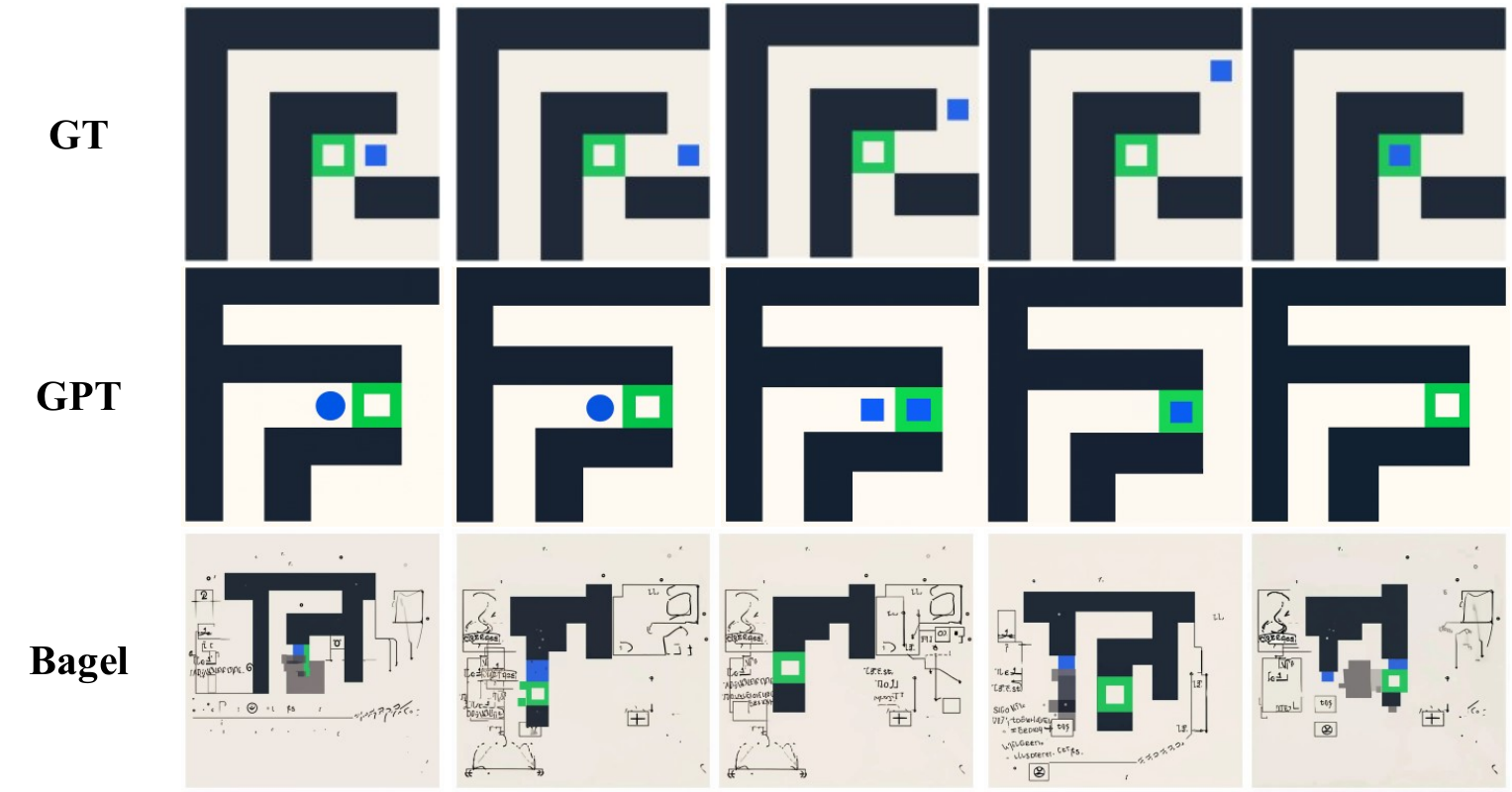}
    \caption{Examples of Maze puzzle.}
    \label{fig:error_maze}
  \end{subfigure}

  \caption{\textbf{ Qualitative case studies illustrating common failure modes of unified models on the Uni-MMMU benchmark.}}
  \label{fig:error_all}
\end{figure}

\subsection{Insights and Analysis}

\noindent\textbf{When is Gen\&Und synergy effective?}
Mutual reinforcement is most beneficial on tasks with a \emph{strict logical dependency} between intermediate states and final outcomes. As evidenced by Tab.~\ref{tab:ab}, coupling the understanding module with intermediate generative states outperforms running understanding and generation in isolation; notably, even \emph{imperfect} intermediates confer gains over the decoupled baseline, while supplying ground-truth intermediates yields consistent and sizeable improvements across all tasks. The effect is especially pronounced for Maze (solution accuracy) and Science (image correctness). As shown in  Fig~\ref{fig:error_svg}, we also observe clear causal handoffs in the \emph{Code Rendering} task: when GPT-4.1 correctly analyzes SVG semantics, GPT-image reliably produces faithful renderings; conversely, misanalysis propagates directly into visual errors. This confirms that a well-defined intermediate pathway is not just beneficial but critical for solving complex, multi-step problems.

\noindent\textbf{Current capability landscape.}
Overall, current unified models exhibit \emph{stronger understanding than generation}, with generation often the bottleneck—consistent with in Tab.~\ref{tab:model_performance} where image-generation scores generally trail understanding scores. As shown in  Fig~\ref{fig:error_all}, common failure modes concentrate in the intermediate steps: (i) instruction-following lapses (adding or omitting requested elements; rendering text-only fields onto images), (ii) background and style drift across edits, (iii) fragile spatial perception and world-knowledge application, and (iv) topology/semantics violations that corrupt downstream reasoning. Concretely, for \emph{Maze}, GPT maintains inter-image consistency yet sometimes distorts wall–path topology or object placement, misleading subsequent planning; Bagel tends to inject extraneous, nonsensical glyphs that render states unparsable. For \emph{Jigsaw}, OmniGen2 frequently copies the $2{\times}2$ reference rather than producing a coherent completion, whereas nano-banana can generate the completion but occasionally introduces irrelevant content that confuses later decisions. Similar background-consistency and instruction-following issues appear in \emph{Sliding Puzzle} and \emph{Geometry}. In \emph{Code}, Ovis-U1 and OmniGen2 often misread SVG semantics (colors, side counts, sizes, or relative positions), while Qwen-Image-Edit erroneously rasterizes the \texttt{Render Summary}—specified as text-only—onto the image. Addressing these deficits will likely require tighter controllability (e.g., program- or constraint-guided generation), stronger spatial/state invariants across edits, and interfaces that make executable intermediate representations first-class citizens in the reasoning–generation loop.

\section{Conclusion}
We introduce Uni-MMMU, a comprehensive benchmark designed to address the gap in evaluating the synergistic capabilities of unified models. Through eight diverse, reasoning-centric tasks, Uni-MMMU assesses models on bidirectionally coupled challenges where either generation aids understanding or understanding aids generation. Our novel evaluation pipeline scores both intermediate processes and final outcomes to enable fine-grained analysis. Our extensive evaluation of state-of-the-art models reveals significant room for improvement, highlighting common failures in precise instruction adherence, and spatial reasoning. To ensure full transparency and reproducibility, we release all code, datasets, evaluation tools, and judge configurations at \url{https://github.com/uni-mmmu/Uni-MMMU}.

\section{Acknowledgements}

This study is supported by the Ministry of Education, Singapore, under its MOE AcRF Tier 2 (MOET2EP20221-0012, MOE-T2EP20223-0002). This research is also supported by cash and in-kind funding from NTU S-Lab and industry partner(s). This work is also supported by Shanghai Artificial Intelligence Laboratory.

\section{Limitations}
While Uni-MMMU takes a significant step toward evaluating the synergistic capabilities of unified multimodal models, we acknowledge several limitations that also highlight promising directions for future research.

First, the scope of tasks in Uni-MMMU is primarily focused on reasoning-centric disciplines with deterministic and verifiable solutions, such as science, coding, and puzzles. This design choice enables objective and reproducible evaluation but does not cover a broader range of real-world scenarios that may require open-ended creativity, subjective judgment, or nuanced commonsense reasoning. Furthermore, the current benchmark is based entirely on static images, leaving the evaluation of models on tasks involving video or longer-term temporal interactions as an area for future work.

Second, our data curation methodologies have inherent trade-offs. For tasks like Maze Navigation, Sliding Puzzle, and Code Rendering, we employed procedural generation to ensure unique solutions and facilitate objective parsing. However, this approach may result in data that lacks the complexity, noise, and visual diversity of real-world imagery. For the Science tasks, we used an LLM-driven pipeline followed by manual curation to ensure scientific validity. While rigorous, this process could potentially introduce subtle biases from the generation models or human curators.

Finally, our evaluation pipeline has its own constraints. For several tasks, we rely on a "model-as-a-judge" framework using powerful VLMs. Although we validated the substantial agreement of our VLM judge with human annotators, these judge models are not infallible and may have their own biases or knowledge gaps that could affect evaluation accuracy.

\bibliography{custom.bib}

\clearpage \newpage
\appendix
\section{Benchmark Details}
\label{sec:appendix}

\subsection{Generation aids Understanding}

\subsubsection{Maze Navigation}

\begin{figure}[h]

\begin{center}
\includegraphics[width=\linewidth]{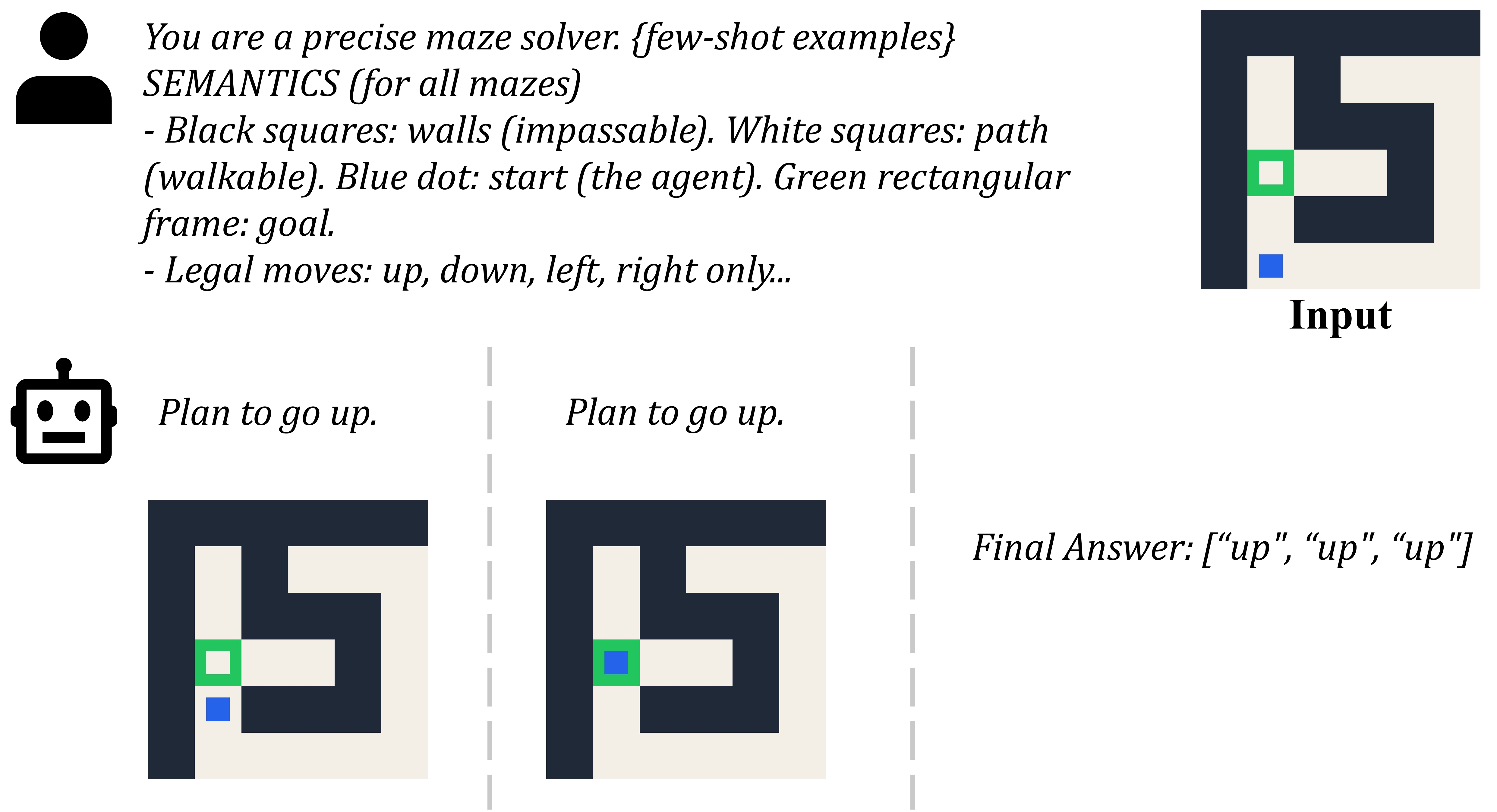}
\end{center}
\caption{\textbf{Details of Maze Puzzle.}}
\label{fig:maze}
\end{figure}

\textbf{Task Definition.}
As shown in Fig~\ref{fig:maze}, given the initial state of a $6 \times 6$ `perfect maze' as an image (blue block for the agent's start position, green frame for the goal, black for walls, and white for paths), the model is tasked with planning and executing the unique shortest path to the goal. The output must be a sequence of state images, each depicting the board after a single move, followed by a final textual representation of the entire path. This task evaluates the model's capabilities in visual parsing, spatial layout comprehension, multi-step spatial reasoning, and adherence to complex output format instructions.

\noindent\textbf{Construction.}
Mazes are procedurally generated to ensure a unique shortest path for unambiguous evaluation. The core algorithm first employs a Depth-First Search (DFS) to carve passages, creating a `perfect maze' with no loops. This guarantees that a single unique path exists between any two cells. The DFS process starts from a random odd-numbered coordinate and uses a step size of two, effectively creating corridors and preventing trivial patterns. Subsequently, a Breadth-First Search (BFS) is used to find the shortest path and verify its uniqueness. Generated mazes are filtered to ensure their shortest path lengths fall within a controlled range of 2 to 10 steps. All visual elements are rendered with a fixed, minimalistic style and color palette to facilitate robust programmatic parsing.

\noindent\textbf{Sampling.}
The prompting strategy is adapted to the model's architecture. For models capable of autonomously generating interleaved image and text outputs, a single prompt is issued to elicit the complete solution. For models that require separate calls for text and image generation, we employ an iterative process: we first prompt the model to generate the text for the next move, then use this plan to prompt the image generator for the corresponding state image. This loop continues until the model outputs the final sequence of path actions.

\noindent\textbf{Evaluation.}
\begin{figure*}[h]
\begin{center}
\includegraphics[width=\textwidth]{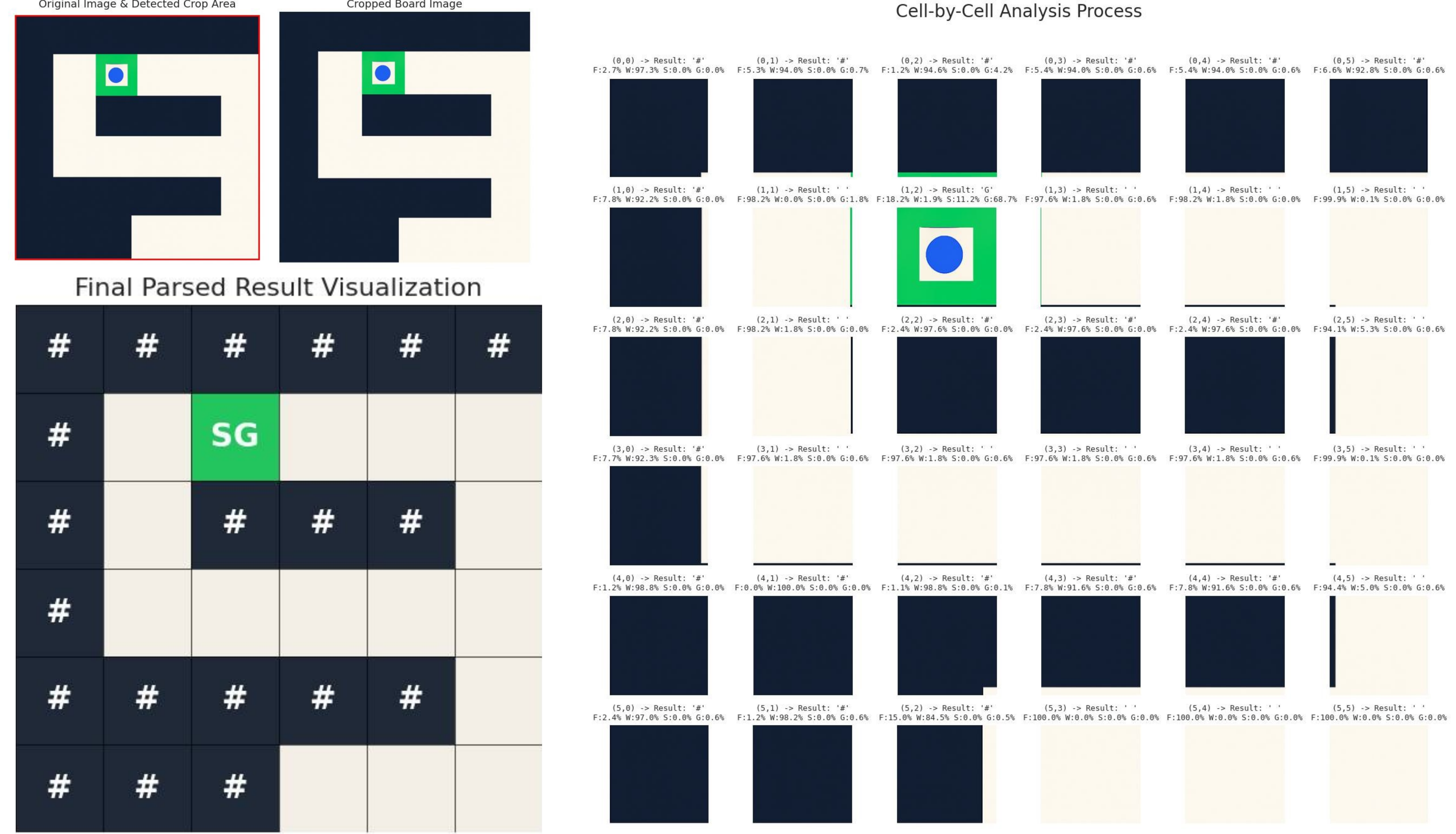}
\end{center}
\caption{\textbf{Overall Pipeline of Programmatic Parser.}}
\label{fig:maze_parse}
\end{figure*}
An automated programmatic parser is used to evaluate each generated image. As shown in Fig~\ref{fig:maze_parse}, the parser first isolates the main maze area from the image, discretizes it into a $6 \times 6$ grid, and converts it into a character-based matrix. This is achieved by analyzing the color distribution of pixels within each cell and calculating their distance to a predefined reference palette. A color tolerance threshold is applied; for instance, a cell is classified as a wall only if over 75\% of its pixels are closer to the wall color than any other color in the palette. Cells that cannot be reliably classified are marked with '?'.

The evaluation metrics are divided into two categories:

\begin{itemize}
    \item \textbf{Intermediate Steps (Image):}
    \begin{itemize}
        \item \texttt{img\_sample\_acc}: A binary score for the entire sample. It is 1 if and only if every generated state image, after parsing, perfectly matches its corresponding ground-truth grid representation.
        \item \texttt{img\_step\_acc}: The fraction of generated images that correctly match their corresponding ground-truth state, calculated over the total number of steps in the ground-truth path.
    \end{itemize}
    \item \textbf{Final Answer (Text):}
    \begin{itemize}
        \item \texttt{text\_sample\_acc}: A binary score. It is 1 if the generated list of moves exactly matches the ground-truth sequence in both content and order.
        \item \texttt{text\_step\_acc}: The proportion of moves in the predicted sequence that correctly match the ground-truth moves, calculated position-wise.
    \end{itemize}
\end{itemize}

\subsubsection{Sliding Puzzle}

\begin{figure}[h]

\begin{center}
\includegraphics[width=\linewidth]{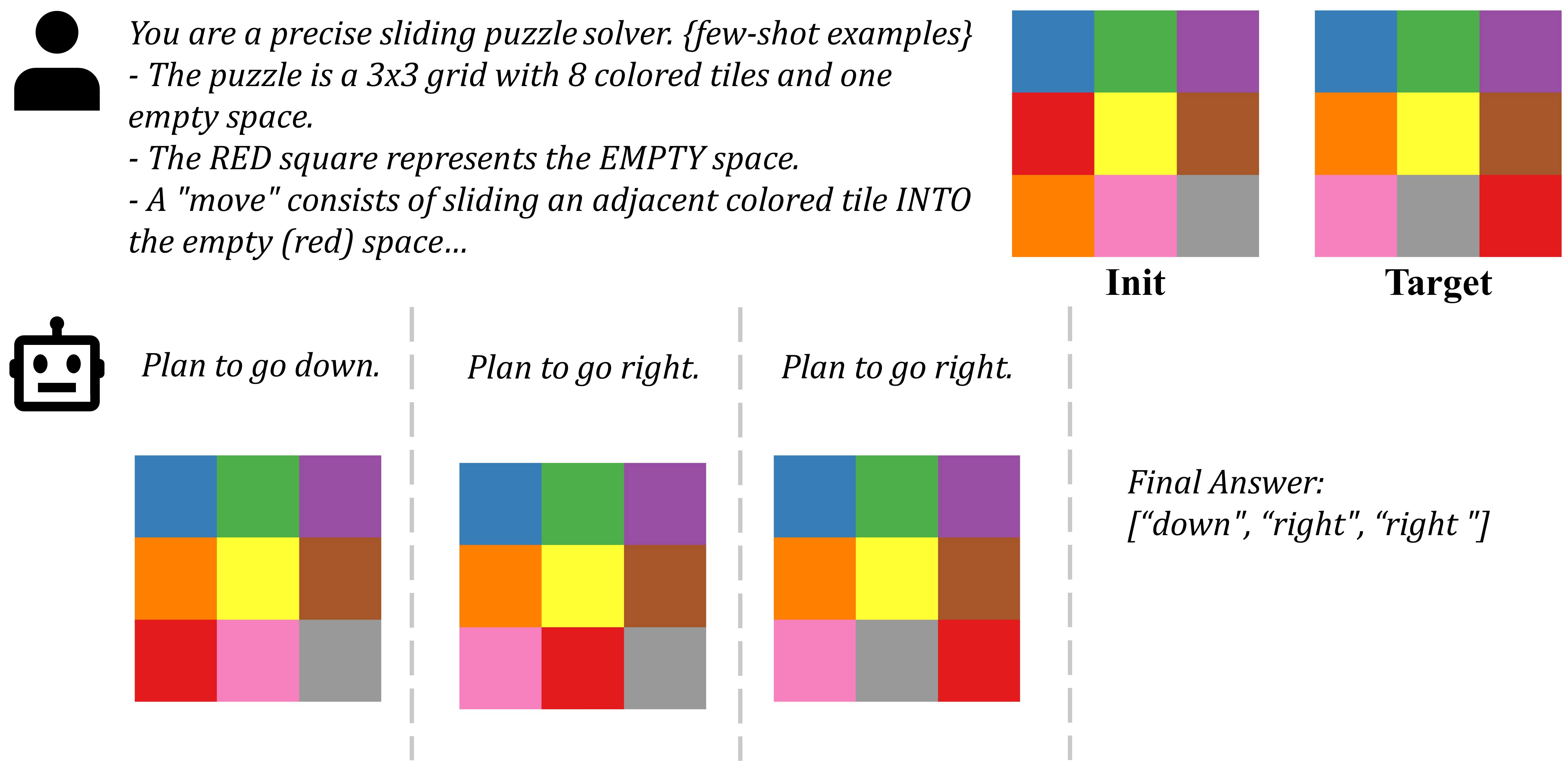}
\end{center}
\caption{\textbf{Details of Sliding Puzzle.}}
\label{fig:sliding}
\end{figure}

\textbf{Task Definition.}
As shown in Fig~\ref{fig:sliding}, given an initial and a final (solved) state of a $3 \times 3$ sliding puzzle (8-puzzle) as images, the model must devise the shortest sequence of moves to solve it. The required output is a series of intermediate state images, each representing the board after one move, followed by a final JSON list of the moves performed. This task assesses a model's ability to parse complex visual states, perform state-space search for optimal planning, and execute the plan through a sequence of generative actions.

\noindent\textbf{Construction.}
Each puzzle instance is procedurally generated to have a unique, optimal solution. The process begins with the solved state, and a number of random moves  are applied to generate a scrambled, yet solvable, initial configuration. A Breadth-First Search (BFS) solver  is then employed to find the length of the shortest solution path and, critically, to count the number of unique shortest paths. An instance is retained only if this count is exactly one, thereby eliminating ambiguity for evaluation. For robust parsing, the nine tiles (8 numbered, 1 empty) are rendered as solid color blocks using a fixed, high-contrast 9-color palette, with no occluding numbers or borders.

\noindent\textbf{Sampling.}
The model is prompted using a multi-modal, few-shot context. This context includes a complete demonstration with example initial/goal states, the sequence of intermediate solution images, and the final JSON answer. As with the Maze task, the inference procedure is adapted to the model's architecture, employing either a single call for models that support autonomous interleaved generation or an iterative, manual prompting process for models requiring separate calls.

\noindent\textbf{Evaluation.}
A programmatic evaluation pipeline assesses the model's output. A dedicated parser  first processes each generated image. It locates the $3 \times 3$ board and discretizes it into a grid of tile identifiers by classifying each cell based on the dominant color's proximity to the reference palette. A strict classification threshold is used: a color must constitute at least 80\% of a cell's pixels (`tolerance=0.80') for the tile to be identified. The final score is based on four metrics, as shown in the Maze task.

\subsubsection{Geometry}

\begin{figure}[h]

\begin{center}
\includegraphics[width=\linewidth]{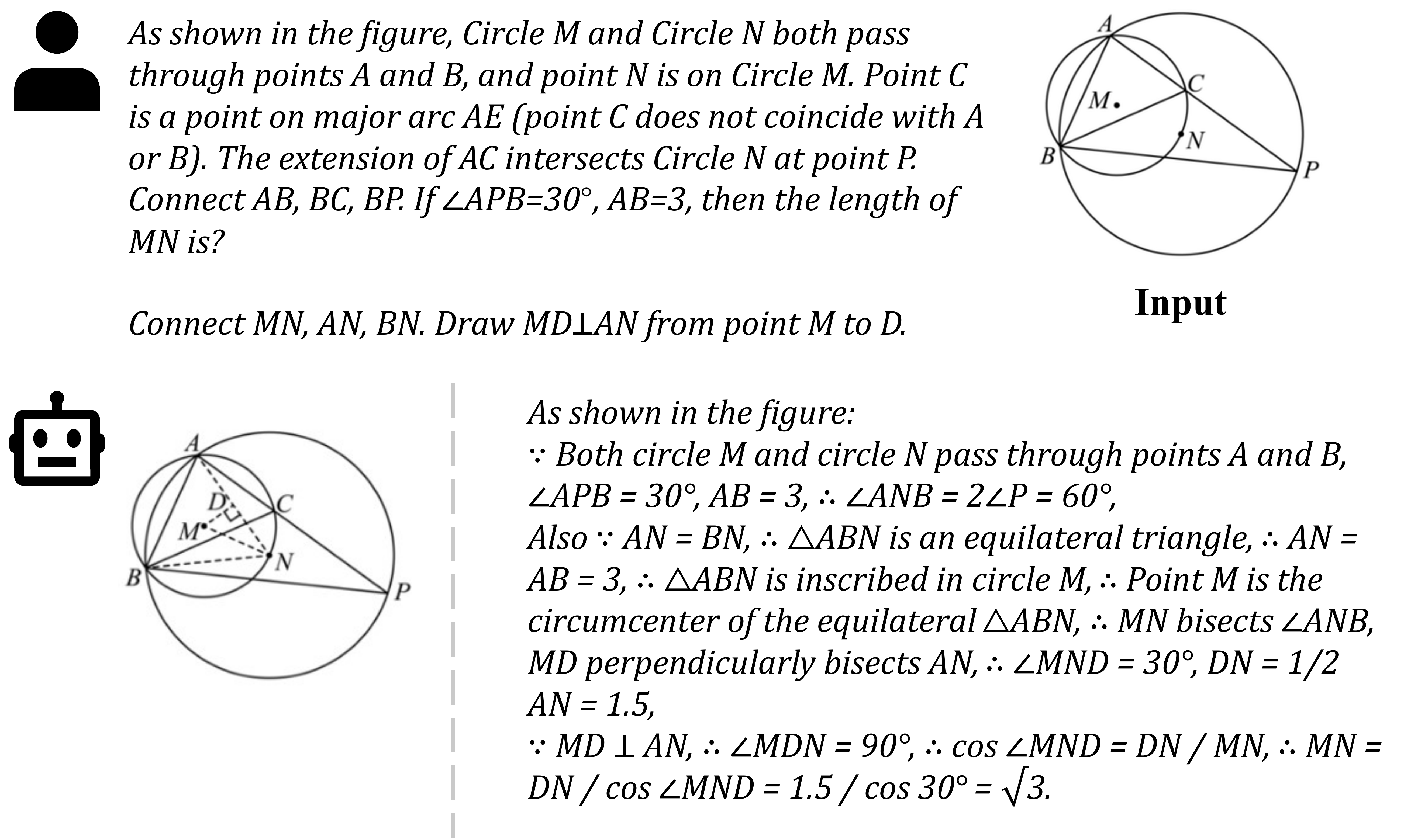}
\end{center}
\caption{\textbf{Geometry.}}
\label{fig:geo}
\end{figure}

\textbf{Task Definition.}
As shown in Fig~\ref{fig:geo}, given a geometry problem consisting of an image and text, the model is required to solve it through a two-stage process that couples generation with reasoning. First, it must interpret textual instructions to generate a new image by accurately drawing specified auxiliary lines on the original figure. Second, using its own generated diagram as a visual aid, it must produce a step-by-step textual solution, which can be either a calculation or a formal proof. This task directly evaluates the "generation aids understanding" paradigm by assessing the model's ability to create meaningful visual constructs to scaffold complex logical deduction.

\noindent\textbf{Construction.}
The dataset comprises 140 challenging problems sampled from the established Geo-Laux benchmark~\cite{fu2025geolaux}. The collection includes a mix of calculation and proof-based questions, each meticulously annotated with a comprehensive set of ground-truth data: (1) the original problem figure, (2) precise textual instructions in English for constructing necessary auxiliary lines, (3) a ground-truth image showing the correctly drawn auxiliary lines, and (4) a complete, step-by-step textual solution and final answer.

\noindent\textbf{Sampling.}
A mandatory two-stage inference process is enforced to ensure the dependency of reasoning on generation. \textbf{Stage 1 (Image Generation):} The model is first prompted with the problem statement, the original figure, and the explicit instructions for the auxiliary lines. It is tasked with performing a precise image editing operation: overlaying these lines and outputting \textit{only the resulting image}. \textbf{Stage 2 (Textual Reasoning):} The image generated by the model in the first stage is then appended to the conversation context. Subsequently, the model is prompted to provide the complete textual solution, compelling it to base its reasoning on the visual cues it has just created.

\noindent\textbf{Evaluation.}
A model-as-a-judge pipeline is employed for a nuanced, dual-component evaluation of the model's output.
\begin{itemize}
    \item \textbf{image\_acc:} The correctness of the generated auxiliary lines is assessed by a powerful Vision-Language Models (VLMs) judge (\texttt{Qwen2.5-VL-72B}). The judge receives the original image, the ground-truth auxiliary line image, the model's generated image, and the textual instructions. It provides a binary score indicating whether all required lines were drawn correctly, with tolerance for minor stylistic variations but not for geometric errors.
    \item \textbf{text\_acc:} The textual solution is evaluated by a separate Large Language Model (LLM) judge (\texttt{Qwen3-32B}). The judge compares the model's reasoning and final answer against the ground-truth solution. It provides a final binary score  which is 1 if and only if both the logical steps are rigorous and the final conclusion (or numerical result) is correct.
\end{itemize}

\subsubsection{Jigsaw}

\begin{figure}[h]

\begin{center}
\includegraphics[width=\linewidth]{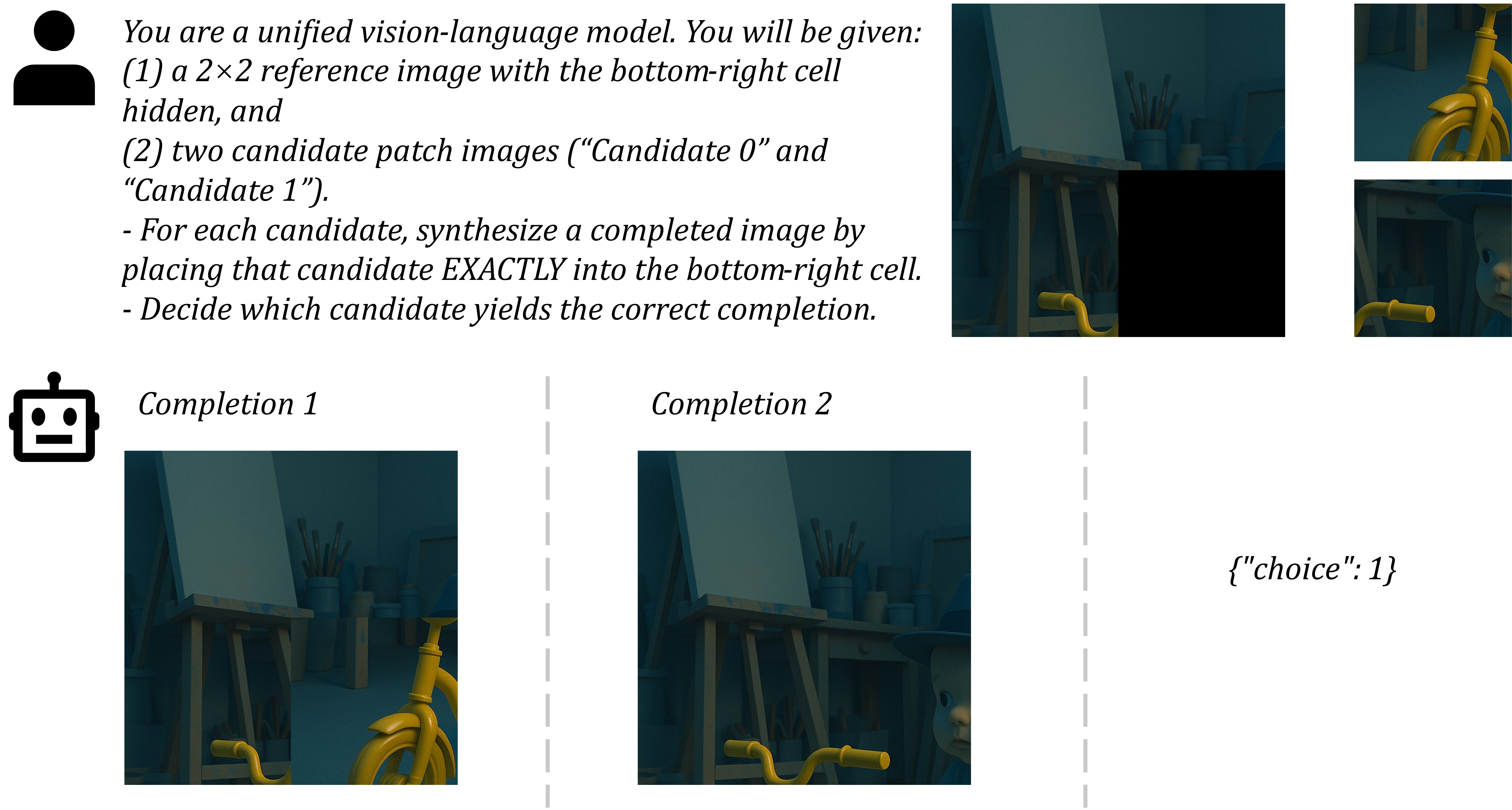}
\end{center}
\caption{\textbf{Details of Jigsaw.}}
\label{fig:jigsaw}
\end{figure}

\textbf{Task Definition.}
As shown in Fig~\ref{fig:jigsaw}, given a $2 \times 2$ image panel with one missing quadrant and two candidate patches, the model must identify the correct patch that completes the image. The task is structured to evaluate both generation and understanding capabilities in sequence. First, the model must generate two composite images, each showing the result of placing one of the candidates into the missing slot. Second, based on its own generated outputs, it must provide a textual analysis and a final decision indicating the correct choice. This task assesses visual reasoning about local and global coherence, including continuity of textures, colors, and geometric structures, as well as the ability to perform conditional image generation and subsequent comparative judgment.

\noindent\textbf{Construction.}
Puzzle instances are generated from a high-quality image dataset~\cite{chen2025sharegpt}. Each source image is first standardized by center-cropping and resizing, then partitioned into a $3 \times 3$ grid of patches. The problem is formulated using the top-left $2 \times 2$ portion of this grid. The patch corresponding to the bottom-right of this $2 \times 2$ panel is designated as the target and is masked out in the reference image provided to the model. The two candidates consist of: (1) the ground-truth target patch, and (2) a distractor patch randomly selected from the remaining patches of the same source image. The positions of the correct and distractor candidates are randomized for each instance.

\noindent\textbf{Sampling.}
The task employs a multi-stage, sequential prompting strategy that compels the model to reason over its own generative outputs. The model is provided with the reference panel and the two candidate patches. The inference proceeds in three steps: \textbf{Generate Completion 1:} The model is prompted to generate the completed $2 \times 2$ image by inserting the first candidate patch into the missing quadrant. \textbf{Generate Completion 2:} The model is then prompted to generate the second completed image using the other candidate patch. \textbf{Analyze and Decide:} With both of its generated composite images included in the context, the model is finally prompted to produce a textual rationale and a structured JSON object containing its final choice (0 or 1).

\noindent\textbf{Evaluation}
The evaluation is twofold, assessing both the quality of the generated images and the accuracy of the final decision.
\begin{itemize}
    \item \textbf{image\_score:} The perceptual similarity of the two generated composite images is measured against their corresponding ground-truth versions using the \textbf{Dream-Sim}~\cite{fu2023dreamsim} metric. A lower Dream-Sim distance indicates a higher quality generation that more accurately reconstructs the scene. A final image score is calculated as \texttt{1.0 - mean\_distance}, with a penalty applied if the model fails to generate exactly two valid images.
    \item \textbf{text\_sample\_acc:} The model's final text output is parsed to extract the chosen candidate index from the structured JSON object. This choice is compared against the ground-truth label to compute a standard classification accuracy.
\end{itemize}
\subsection{Understanding aids Generation}

\subsubsection{Science}

\begin{figure}[h]

\begin{center}
\includegraphics[width=\linewidth]{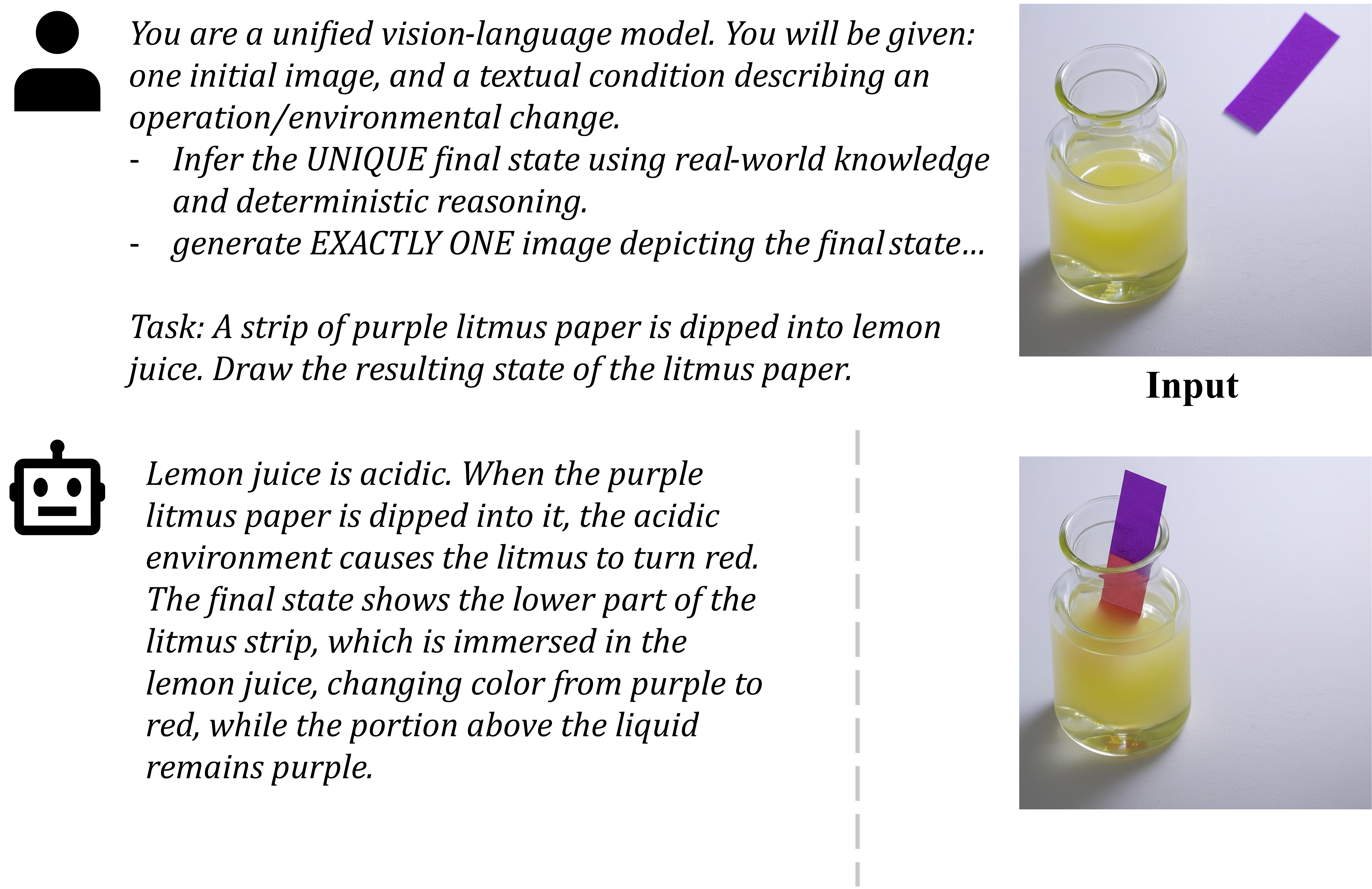}
\end{center}
\caption{\textbf{Details of Natural sciences(physics, chemistry, and biology) Tasks}}
\label{fig:sci}
\end{figure}

\textbf{Task Definition.}
As shown in Fig~\ref{fig:sci}, this task assesses a model's ability to apply fundamental principles from natural sciences (physics, chemistry, and biology) to predict and visualize the outcome of physical processes. Given an image depicting an initial state and a textual description of a change or condition, the model must first provide a textual explanation of the resulting final state based on scientific reasoning. Subsequently, it must generate an image that visually represents this final state. This "understanding aids generation" task evaluates the model's world knowledge and its capacity to use causal reasoning to guide a generative process.

\noindent\textbf{Construction.}
The dataset is constructed through a hierarchical, LLM-driven pipeline. A large language model (GPT) is prompted to generate a wide range of scientific scenarios, starting from broad categories (e.g., Thermal, Fluid Mechanics, Chemistry, Biology) and refining them into specific principles (e.g., Thermal Expansion, Oxidation, Phototropism). Within each scenario, it produces a structured set of text-to-image prompts, including a detailed initial image prompt to create the starting scene and an output image editing prompt describing the transformation to the final state. These prompts are then fed into a unified generation and editing model (nano-banana) to synthesize the initial and ground-truth final images. Every resulting image pair undergoes a rigorous manual curation process, where human experts verify its scientific accuracy, visual plausibility, and ensure the outcome is unambiguous and deterministic. Samples that do not meet these standards are either regenerated or discarded entirely to maintain the integrity of the benchmark.

\noindent\textbf{Inference Procedure.}
A two-stage "reason, then generate" inference process is enforced. \textbf{Stage 1 (Scientific Reasoning):} The model is provided with the initial state image and a text prompt describing the applied condition (e.g., "The temperature is raised to 100 degrees Celsius," "A zinc strip is placed into the solution"). It is first required to output a textual explanation (\texttt{<OUTPUT\_PROMPT>}) detailing the scientific principles at play and describing the resulting final state. \textbf{Stage 2 (Image Generation):} The model's own textual reasoning from the first stage is then added to the conversation context. Subsequently, the model is prompted to generate a single image that visually depicts the final state it just described.

\noindent\textbf{Evaluation.}
The evaluation is conducted using a dual-component, model-as-a-judge pipeline, assessing both the textual reasoning and the generated image.
\begin{itemize}
    \item \textbf{text\_reason\_acc:} A Vision-Language Models (VLMs) judge (\texttt{Qwen2.5-VL}) evaluates the model's textual output a binary score indicating whether the explanation correctly applies relevant scientific principles.
    \item \textbf{text\_result\_acc:} A binary score indicating whether the described final state is physically plausible and accurate.
    \item \textbf{img\_acc:} The VLMs judge provides a binary score (\texttt{image\_correct}) assessing whether the image semantically matches the expected outcome while maintaining the consistency of the initial scene.

\end{itemize}

\subsubsection{Code}

\begin{figure}[h]
\begin{center}
\includegraphics[width=\linewidth]{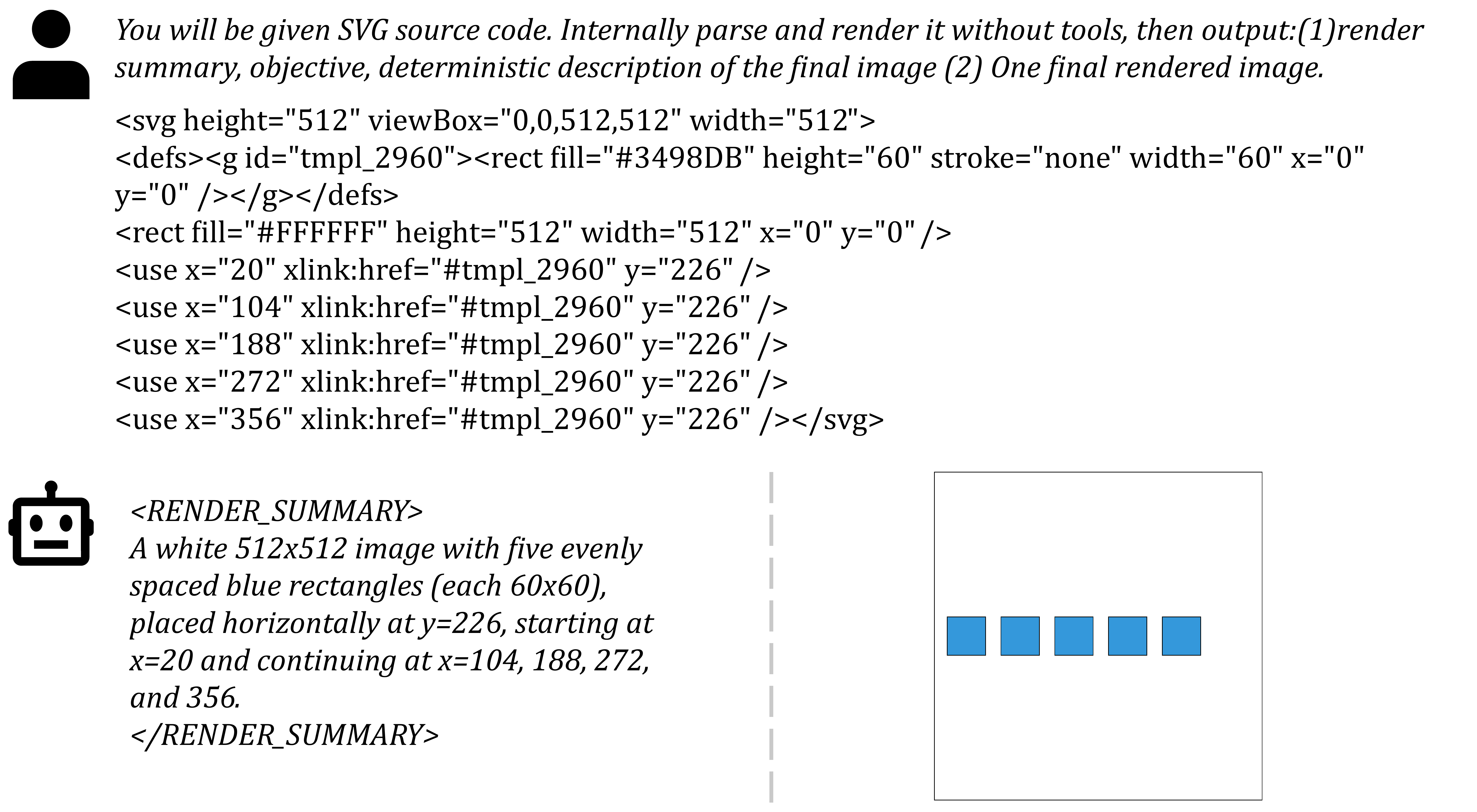}
\end{center}
\caption{\textbf{Details of Code Rendering.}}
\label{fig:svg}
\end{figure}

\textbf{Task Definition.}
As shown in Fig~\ref{fig:svg}, this task evaluates a model's ability to interpret and render Scalable Vector Graphics (SVG) source code without an external interpreter. Given raw SVG code as input, the model must first generate a textual description summarizing the visual output it intends to create. It then must generate the final rendered image. This "understanding aids generation" task probes the model's intrinsic knowledge of the SVG specification, its capacity to parse declarative code, and its ability to translate programmatic logic—including control flow—into a precise visual representation.

\noindent\textbf{Construction.}
The dataset is procedurally generated with three tiers of escalating difficulty to test a range of capabilities.
\begin{itemize}
    \item \textbf{Simple:} A single geometric primitive (e.g., circle, rectangle, or polygon with 3-6 sides).
    \item \textbf{Medium:} Multiple non-overlapping shapes or the introduction of simple control flow.
    \item \textbf{Complex:} Multiple potentially overlapping shapes, including lines and curves, and more advanced control flow.
\end{itemize}
A crucial feature is the inclusion of programmatic control flow, which is synthesized in the SVG code through constructs like `for' loops (emulated as regularly spaced, repeated elements) and `<defs>'+`<use>' blocks (defining a template and instantiating it multiple times). All visual elements are drawn from a fixed 8-color palette. 

\noindent\textbf{Sampling.}
A two-stage "understand, then generate" inference process is mandated. \textbf{Stage 1 (Code Understanding):} The model receives the raw SVG code and is first prompted to output a concise textual summary (\texttt{<RENDER\_SUMMARY>}) of the visual scene the code describes. \textbf{Stage 2 (Image Generation):} The model's own generated summary from Stage 1 is then added to the context, and it is subsequently prompted to render the final image. This sequence encourages the model to leverage its textual interpretation to guide the visual synthesis.

\noindent\textbf{Evaluation.}
A Vision-Language Models (VLMs) judge (\texttt{Qwen2.5-VL}) performs a qualitative, multi-faceted evaluation of the model's outputs against the ground-truth rendered image.
\begin{itemize}
    \item \textbf{text\_acc:} The VLMs judge compares the model-generated summary against the ground-truth image to assess semantic consistency, checking for the correct object types, counts, colors, and relative layout. This yields a binary accuracy score.
    \item \textbf{shape\_color\_acc:} The VLMs judge evaluates the model-generated image using a detailed rubric, providing scores on a 0-5 scale for two distinct axes: Assesses the correctness of object types (e.g., circle vs. polygon), side counts for polygons, and adherence to the specified color palette. 
    \item\textbf{position\_acc:} use VLMs as Judge; Assesses the correctness of the overall layout, including relative positions, alignment, spacing, layering (z-order), and rotation.
\end{itemize}

\section{Elaboration}
\noindent\textbf{Potential Risks.} The work focuses on creating a benchmark for evaluating AI capabilities on reasoning tasks like puzzles and scientific problems. The potential risks are minimal and indirect, related to the general misuse of advanced AI, which is outside the scope of this specific benchmark's contribution.

\noindent\textbf{Budget and Setups.} All experiments were conducted on NVIDIA A800 GPUs, consuming approximately 48 GPU-days in total. Sampling parameters for all models were set to their official default values.

\section{Quantitative Failure-Mode Analysis}
\label{sec:failure_quant}

To complement the qualitative case studies in Fig.~\ref{fig:error_all}, we provide per-task quantitative statistics for the most frequent failure modes across unified models:

\begin{itemize}[leftmargin=10pt,topsep=2pt]
    \item \textbf{Jigsaw.} nano-banana generates an incorrect number of images in 13.3\% of cases and produces irrelevant images (DreamSim distance $< 0.2$) in 41.6\%; OmniGen2 copies the $2{\times}2$ reference panel instead of producing a coherent completion in 100\% of cases.
    \item \textbf{Code Rendering.} Qwen-Image-Edit erroneously rasterizes the text-only Render Summary onto the generated image in 55\% of cases; Ovis-U1 does so in 100\% of cases. Both models also frequently misread SVG semantics (wrong colors, polygon side counts, or relative positions).
    \item \textbf{Maze.} Invalid images—those containing any grid cell that cannot be parsed into a valid category (floor, wall, start, or goal)—occur for 15\% of nano-banana outputs, 84\% of GPT outputs, and 100\% of Bagel outputs.
    \item \textbf{Sliding Puzzle \& Geometry.} The dominant failures are background/style drift across edits and instruction-following lapses (omitting requested auxiliary lines or adding extraneous elements), which together account for the majority of score degradation.
\end{itemize}

These statistics confirm that the primary bottleneck lies in \emph{visual generation fidelity}—especially precise spatial editing and instruction adherence—rather than in textual reasoning.

\end{document}